\documentclass[cameraready]{Interspeech}
\title{Beyond Binary: Speech Representations Across the Cognitive Score Hierarchy}

\author[affiliation={1,2}, orcid=0000-0002-1409-2896 ]{Serli}{Kopar} %
\author[affiliation={1,2,3},orcid=0000-0002-3996-2034]{Roshan Prakash}{Rane}
\author[affiliation={4,5}, orcid=0000-0002-9234-6046]{Christian}{Mychajliw}
\author[affiliation={1,2}, orcid=0009-0006-4500-2175]{Lydia}{Federmann}

\author[affiliation={4,5,6}, orcid=0009-0007-2630-5928]{Gerhard}{Eschweiler}
\author[affiliation={7,8}, orcid=0000-0001-5796-5442]{Daniela}{Berg}
\author[affiliation={1,2,11}, orcid=0000-0003-3275-9022]{Sam}{Gijsen}
\author[affiliation={9}, orcid=0000-0002-2727-2116]{Paula Andrea}{Perez-Toro}
\author[affiliation={1,2,10}, orcid=0000-0001-7115-0020]{Kerstin}{Ritter}
\address{
\normalsize
    $^1$ Hertie Institute for AI in Brain Health, University of Tübingen, Tübingen, Germany
    $^2$ Tübingen AI Center, University of Tübingen, Tübingen, Germany 
    $^3$ Department of Psychology, Humboldt-Universität zu Berlin 
    $^4$ Geriatric Center, Tübingen University Hospital, Tübingen, Germany 
    $^5$ Tübingen Center for Mental Health (TüCMH), Department of Psychiatry and Psychotherapy, Tübingen University Hospital, Tübingen, Germany 
    $^6$ German Center for Mental Health (DZPG), Partner Site Tübingen, Tübingen, Germany 
    $^7$ Department of Neurology, University Medical Center Schleswig-Holstein and Kiel University, Kiel, Germany 
    $^8$ Center for Neurology, University Hospital Tübingen and Hertie Institute for Clinical Brain Research, Tübingen, Germany
    $^9$Pattern Recognition Lab, Friedrich-Alexander-Universität Erlangen-Nürnberg, Erlangen, Germany 
    $^{10}$  Charit\'e--Universit\"atsmedizin, Department of Psychiatry and Psychotherapy, Berlin, Germany 
}

\email{serli.kopar@uni-tuebingen.de,  paula.andrea.perez@fau.de, kerstin.ritter@uni-tuebingen.de}


\usepackage{comment}
\usepackage{booktabs} % For professional lines
\usepackage{multirow}
\usepackage[table]{xcolor}
\usepackage{graphicx}
\usetikzlibrary{shapes.geometric, arrows.meta, positioning}
\usepackage{multirow}
\usepackage{float} 
\usepackage{pdflscape}
\usepackage{tikz}
\usepackage{svg}
\usepackage[english]{babel}

\begin{document}

\maketitle


\begin{abstract}
This study examines the relationship between speech representations and the hierarchical structure of cognitive assessment in mild cognitive impairment. Utilizing 5,754 German neuropsychological assessment recordings, we evaluate six cognitive tasks across three score levels: task, domain, and global levels. We compare hand-crafted acoustic features with self-supervised learning (SSL) embeddings. Results show that although SSL representations generally outperform hand-crafted features at lower levels, this trend reverses for MCI classification. Furthermore, task-specific constraints influence performance: tasks with greater response freedom exhibit performance dilution as hierarchical levels increase, suggesting ``specialist'' representations, whereas the performance of highly structured tasks increases toward higher levels, suggesting ``generalist'' representations. These findings show links between task constraints and assessment hierarchy in automated clinical speech analysis.
\end{abstract}

\section{Introduction}
Mild cognitive impairment (MCI) is a clinical syndrome characterized by cognitive decline exceeding normal aging \cite{Portet714, Smid2022Subjective}. As a prodromal stage of dementia, most commonly Alzheimer’s disease, it represents a critical window for early intervention \cite{Anderson_2019}. Despite its clinical relevance, MCI remains substantially underdiagnosed \cite{Bohlken2019MCI}. Clinical diagnosis typically relies on standardized neuropsychological assessments. The Consortium to Establish a Registry for Alzheimer’s disease (CERAD+) is a well-validated, multi-domain neuropsychological battery assessing language, memory, executive function, and visuospatial abilities \cite{morris1989cerad}, generating structured task-, domain-, and global-level scores. In routine practice, shorter instruments such as the Mini-Mental State Examination (MMSE) are frequently used for efficient screening. Together, these instruments form the backbone of clinical cognitive assessment. Automated speech analysis has emerged as a promising complement to traditional testing \cite{Konig2015Automatic}. However, current approaches face three methodological bottlenecks: (i) focusing on binary classification (Alzheimer's Disease vs.\ healthy controls (HC)), which lacks sensitivity to the subtle cognitive change characteristics of MCI~\cite{Mekulu2025}; (ii) reliance on English-centric, single-task datasets, limiting generalizability~\cite{Ding2024Speech}; and (iii) modeling clinical scores as independent, flat targets, thereby ignoring the hierarchical structure inherent to standardized cognitive assessment.

In this paper, we address these gaps using 5,754 recordings collected during five CERAD+ tasks and one MMSE screening task from an elderly German cohort. Beyond binary classification, we model the hierarchical organization of clinical scores, linking acoustic features to task-, domain-, and global-level scores. This enables a fine-grained analysis of the relationship between speech and cognitive decline across multiple diagnostic and screening tasks. Our analysis reveals a task-dependent pattern: for tasks with more open-ended responses, the predictive power of acoustic features decreases from task-level to domain-level and global scores, whereas for more constrained tasks, it increases across these levels. To the best of our knowledge, this is the first study to examine how acoustic feature predictiveness varies across hierarchical levels of the German CERAD+ battery in the context of MCI. To support future research, we publicly release our code.\footnote{https://github.com/anon-interspeech/anon-interspeech-2026.git}
%Reviewer Comment: UPLOAD AS ZIP OR ANONYMOUS REPO

\begin{figure*}[t]
  \centering
  \includegraphics[width=0.95\linewidth]{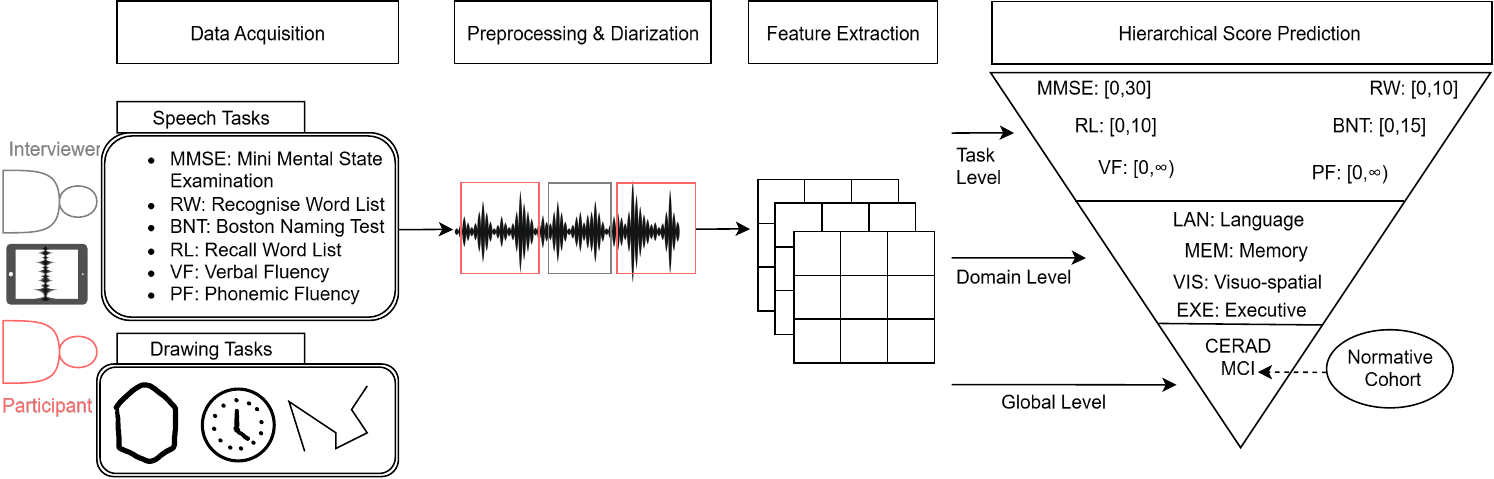}
  \caption{Our workflow for predicting hierarchical cognitive scores from speech. Using speech-derived acoustic features, independent models predict targets at three levels:
(1) \textbf{Level 1 (Individual Tests)}: task-level scores (e.g., Phonemic Fluency (PF), the number of valid words beginning with “S”, with scores ranging from $[0, \infty)$);
(2) \textbf{Level 2 (Cognitive Domains)}: domain-level composite scores with speech-to-drawing task ratios of 1:0 (Language, LAN), 1:1 (Memory, MEM), and 0:1 (Executive, EXE; Visuospatial, VIS) \cite{Roberts2008Mayo}; and
(3) \textbf{Level 3 (Global Status)}: global-level scores including the CERAD+ total score (modeled as continuous and binary with a threshold of 85) \cite{Chandler2005} and binary MCI status (defined as more than 1.5 standard deviations below the normative cohort) \cite{Berres2000}. Arrows denote workflow from shared acoustic feature representations to independent models predicting hierarchical targets.}
  \label{fig:speech_production}
\end{figure*}

\section{Methods}

\subsection{Dataset and Quality Control}
\label{subsec:dataset}
We used speech recordings from the TREND study\footnote{All participants provided written informed consent. The study was approved by the Ethics Committee of (anonymized)} \cite{trendstudie2026}, comprising one MMSE screening task and five CERAD+ diagnostic tasks: Word List Recognition (RW), Boston Naming Test (BNT), Word List Recall (RL), Verbal Fluency (VF), and Phonemic Fluency (PF). Our analysis follows the inherent three-level hierarchy of clinical assessment, as summarized in Fig.~\ref{fig:speech_production}:
\textbf{Level~1 (Individual Tests)} comprises raw task-level scores (e.g., PF: number of valid words beginning with “S”).
\textbf{Level~2 (Cognitive Domains)} aggregates these task-level scores into domain-level composite measures~\cite{Roberts2008Mayo}, including Language (LAN), Memory (MEM), Executive Function (EXE), and Visuospatial Ability (VIS). These domains differ in their reliance on verbal versus drawing-based tasks (LAN: 1:0; MEM: 1:1; EXE and VIS: 0:1).
Although EXE and VIS are traditionally assessed through drawing-based tasks, we evaluate the cross-domain predictive performance of speech by using acoustic features from verbal tasks to predict these non-verbal targets. This tests whether speech contains information that generalizes beyond verbal cognitive domains (MEM, LAN). \textbf{Level~3 (Global Status)} includes the global-level scores: CERAD+ total score (continuous and thresholded at 85)~\cite{Chandler2005} and clinical MCI status ($>1.5$~SD below demographically adjusted normative cohort) reported by Berres et al.~\cite{Berres2000}.\\
To ensure diagnostic and acoustic integrity, we excluded non-native speakers, incomplete profiles, and MCI-to-HC reverters. Additionally, we performed acoustic quality assessment, enforcing constraints on duration (minimum 15~s), energy (RMS $>-55$~dBFS), digital clipping ($<1.5\%$), and signal-to-noise ratio (SNR $>10$~dB), estimated using reference-free, quantile-based methods~\cite{NISTHist,SNREst,ProsodyCog}. Conditional inconsistencies between metrics (e.g., high speech activity ratio with low SNR) were manually reviewed. These filtering steps yielded 959 sessions (698 HC, 261 MCI) from 593 participants.

\subsection{Preprocessing and Diarization Optimization}
\label{subsec:preprocessing}
To find the optimal preprocessing hyperparameters, we used a manually transcribed ground-truth subset ($N=89$; $\approx9\%$ of the corpus, available only in two fluency tests: Phonemic (PF) and Verbal (VF)). We conducted a grid search of $>2{,}500$ combinations using participant-disjoint tuning and validation splits. Hyperparameter tuning was based on diarization and Jaccard error rates (DER, JER), purity (PUR), and coverage (COV), with a 250~ms collar~\cite{pyannote_metrics}. The resulting pipeline applied a 6th-order Butterworth high-pass filter ($f_c = 100$~Hz)~\cite{butterworth1930filter}, spectral-gating noise suppression ($\alpha = 0.3$)~\cite{1163209}, and loudness normalization ($-23$~LUFS)~\cite{pyloudnorm}. The final configuration achieved 0.20 DER, 0.33 JER, 94\% PUR and 97\% COV on the participant-disjoint validation split.
To enable high-density voice-quality features, we generated two audio streams: \textit{Prosody-Preserved} (examiner masked, temporal structure retained) and \textit{Concatenated} (participant segments merged using 10~ms linear cross-fades). Both streams were manually audited to ensure transition integrity. 

\subsection{Feature Extraction}
From these streams, we extracted the extended Geneva Minimalistic Acoustic Parameter Set (eGeMAPS)~\cite{Eyben16}. We computed prosodic features (EG Prosody) from the \textit{Prosody-Preserved} stream to retain conversational timing and voice-quality features (EG V-Qual) from the \textit{Concatenated} stream to enable high-density voice-quality features. We combined them into a third, unified \textit{EG All} feature set. In addition, following recent work on dementia detection with semantic and phonemic fluency tasks \cite{HuBERT_Dyasrthia, SAPKOTA2025103326, HUBERT_last_layer}, we extract latent representations from the frozen final hidden layers of \textit{wav2vec~2.0} (W2V2; \texttt{facebook/wav2vec2-base-960h}) and \textit{HuBERT} (\texttt{facebook/hubert-large-ls960-ft}) using global mean pooling from \textit{Prosody-Preserved} stream. To validate our findings in an independent participant sample, we split the dataset into subject-disjoint \textit{development} and \textit{hold-out} sets. We then confirmed that these sets were comparable (Tab.~\ref{tab:dataset_statistics_exp_hold}) within both HC and MCI by testing differences in hierarchical scores, age and sex (chi-square ($\chi^2$) tests for categorical variables and $t$-tests for continuous variables). No significant differences were observed ($p>0.05$).

\begin{table}[!htpb]
\caption{Demographic statistics and cognitive assessment scores for the development and hold-out sets. Values are presented as Mean ($\pm$ standard deviation (SD)) or Count (\%).}
\label{tab:dataset_statistics_exp_hold}
\centering
\scriptsize
\setlength{\tabcolsep}{3pt}
\renewcommand{\arraystretch}{1.0}

\begin{tabular}{lcccc}
\toprule
 & \multicolumn{2}{c}{\textbf{Development (N=772)\textsuperscript{1}}} & \multicolumn{2}{c}{\textbf{Hold-out (N=187)}} \\
\cmidrule(lr){2-3} \cmidrule(lr){4-5}
\textbf{Feature} & \textbf{HC} & \textbf{MCI} & \textbf{HC} & \textbf{MCI} \\
\midrule
Subjects (N) & 359 & 115 & 88 & 31 \\
Age (years)  & 73.1 $\pm$ 6.1 & 74.9 $\pm$ 5.8 & 73.0 $\pm$ 6.0 & 74.9 $\pm$ 6.8 \\
Sex (\% Female) & 53.2\% & 36.5\% & 55.7\% & 35.5\% \\
Phonemic Fluency (PF) & 14.8 $\pm$ 4.6 & 13.1 $\pm$ 4.8 & 15.0 $\pm$ 4.2 & 13.3 $\pm$ 4.6 \\
MMSE & 28.3 $\pm$ 1.5 & 27.6 $\pm$ 2.0 & 28.3 $\pm$ 1.4 & 27.6 $\pm$ 2.0 \\
Language Domain (LAN) & 0.78 $\pm$ 0.1 & 0.74 $\pm$ 0.1 & 0.78 $\pm$ 0.1 & 0.72 $\pm$ 0.1 \\
CERAD+ (Total) & 85.7 $\pm$ 8.2 & 79.5 $\pm$ 10.6 & 85.3 $\pm$ 8.1 & 77.5 $\pm$ 11.6 \\
Avg. Rec.\textsuperscript{2} & 1.7 $\pm$ 0.6 & 1.4 $\pm$ 0.5 & 1.6 $\pm$ 0.6 & 1.5 $\pm$ 0.6 \\
\bottomrule
\end{tabular}

\vspace{3pt}
\textsuperscript{1} Number of recordings. \textsuperscript{2} Average recordings per participant.
\end{table}

\subsection{Prediction and Validation Framework}
To evaluate prediction performance on the \textit{development set}, we employed $5\times3$ nested cross-validation (NCV). For each hierarchical target, task, and feature set, models were trained from scratch with strictly subject-disjoint folds, ensuring that no participant appeared in multiple folds or across training and validation/test sets. The pipeline incorporated $z$-score normalization and PCA variance thresholding (including a \textit{passthrough} option) within the inner-loop grid search. We evaluated Ridge regression, support vector machines (SVM for classification; SVR for regression), and extreme gradient boosting (XGBoost). Inner-loop optimization targeted balanced accuracy for classification and $R^2$ for regression. After NCV, the best-performing model architecture for each target was selected based on mean outer-fold performance, and only these results are reported for the development set. To determine the final configuration of model hyperparameters to be tested on a disjoint hold-out set, we used a majority vote across NCV folds. This model was then retrained from scratch on the full development set and evaluated on the hold-out set to verify generalization to unseen participants. All models were implemented in Python using \texttt{scikit-learn}~(v1.8.0)~\cite{JMLR:v12:pedregosa11a} and \texttt{xgboost}~(v3.1.2)~\cite{Chen2016}.

\section{Results}

\subsection{Level~1: Task-Level Score Prediction}
\begin{figure}[!htpb]
    \centering
    \includegraphics[width=\columnwidth]{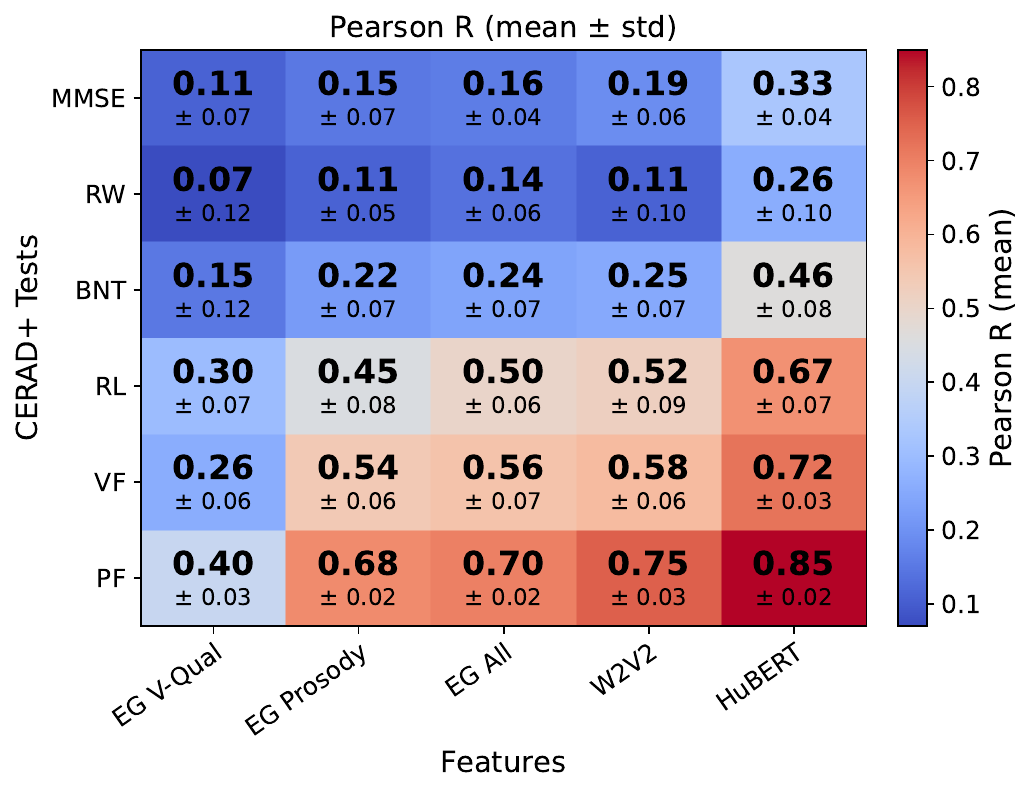}
    \caption{Level 1: Individual test score prediction. Pearson correlation ($r$) between predicted and ground-truth scores. The x-axis shows feature sets; the y-axis lists MMSE and CERAD+ subtests ordered by increasing response freedom (from constrained to open-ended). Values denote mean $\pm$ SD across cross-validation folds.} %Abbreviations: MMSE: Mini-Mental State Examination; RW: Word List Recognition; BNT: Boston Naming Test; RL: Word List Recall; VF: Verbal Fluency; PF: Phonemic Fluency; EG-VQual: eGeMAPS voice-quality; EG-Prosody: eGeMAPS prosodic features; EG-All: full eGeMAPS feature set; W2V2: wav2vec~2.0.}
    \label{fig:level1_test_match}
\end{figure}

Level~1 results (Fig.~\ref{fig:level1_test_match}) report mean ($\pm$ SD) Pearson correlations on the development set for predicting individual task-level scores using features extracted from single neuropsychological tasks. Two consistent trends emerge. First, performance improves from hand-crafted eGeMAPS features (EG V-Qual) to self-supervised learning (SSL) representations across all assessments, with HuBERT yielding the highest prediction performance. Second, performance increases as tasks allow more open-ended responses: constrained tasks (MMSE, RW, BNT) show weaker performance, whereas open-ended tasks (VF, PF) show stronger performance.

\subsection{Level~2: Cognitive Domain Score Predictions}
\begin{figure*}[!htpb]
    \centering
    \includegraphics[width=0.82\linewidth]{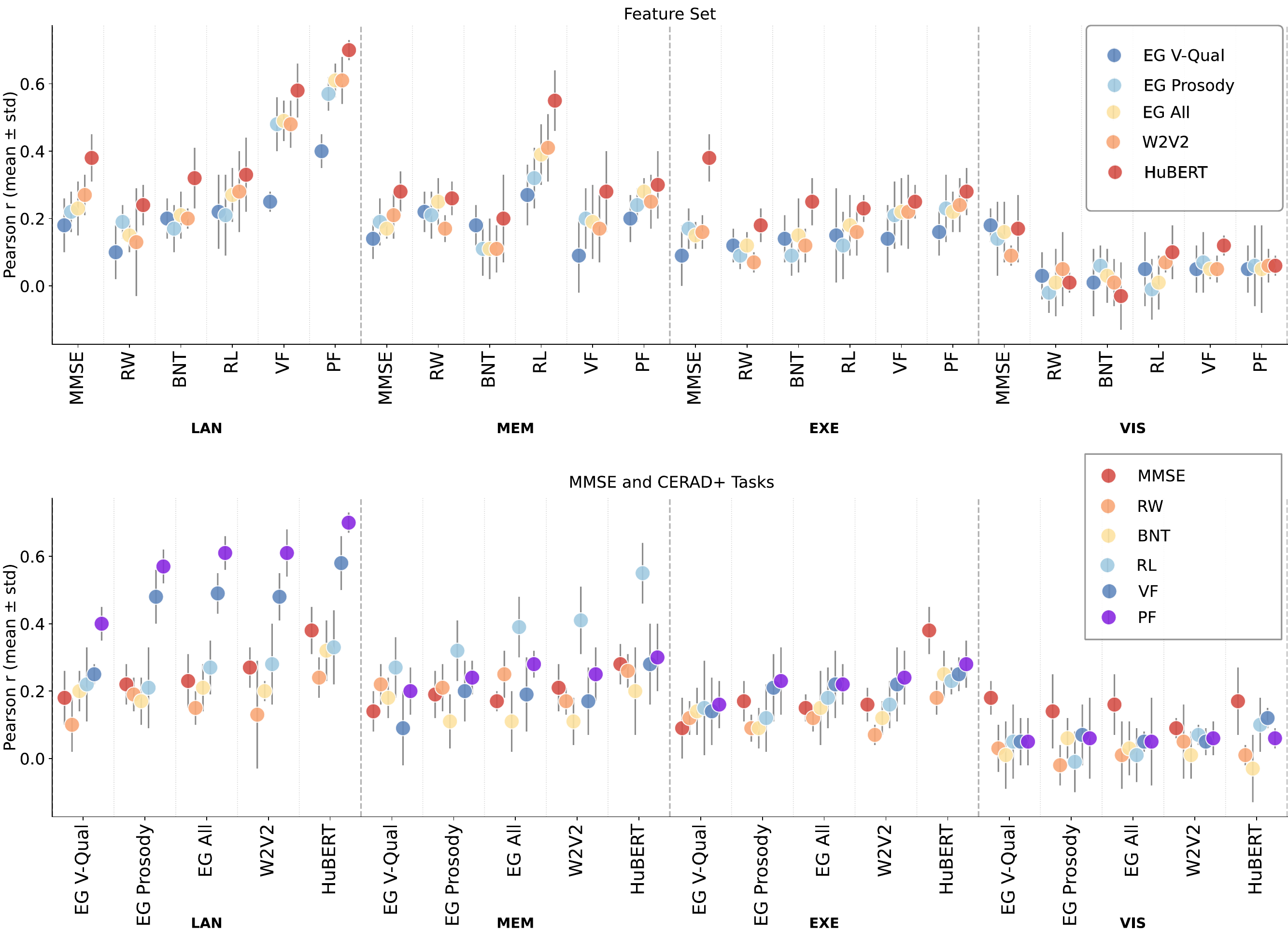}
    \caption{Level~2: Cognitive domain score prediction (LAN: Language, MEM: Memory, EXE: Executive Function, VIS: Visuospatial Ability). The upper panel shows mean ($\pm$ SD) Pearson correlations by feature set across tasks, and the lower panel shows results by input task across feature sets. Tasks are ordered within domains by increasing degrees of freedom, from constrained to open-ended.}
    \label{fig:level2_alt}
\end{figure*}

Moving from individual tests to domains, Level~2 results (Fig.~\ref{fig:level2_alt}) evaluate prediction of composite domain-level cognitive scores (LAN, MEM, EXE, VIS). Performance is analyzed across feature sets (upper panel) and input tasks (lower panel). In the upper panel, HuBERT consistently achieves the strongest performance, while eGeMAPS All remains competitive and often matches W2V2. Performance drops in the drawing-based EXE and VIS domains. In the lower panel, task-level analysis shows that PF and VF are the strongest predictors within LAN. Within MEM, RL emerges as the dominant task. Notably, MMSE performs equally well for EXE and LAN ($r = 0.38$), followed by MEM and VIS.

\begin{figure}[t]
  \centering
  \includegraphics[width=\linewidth]{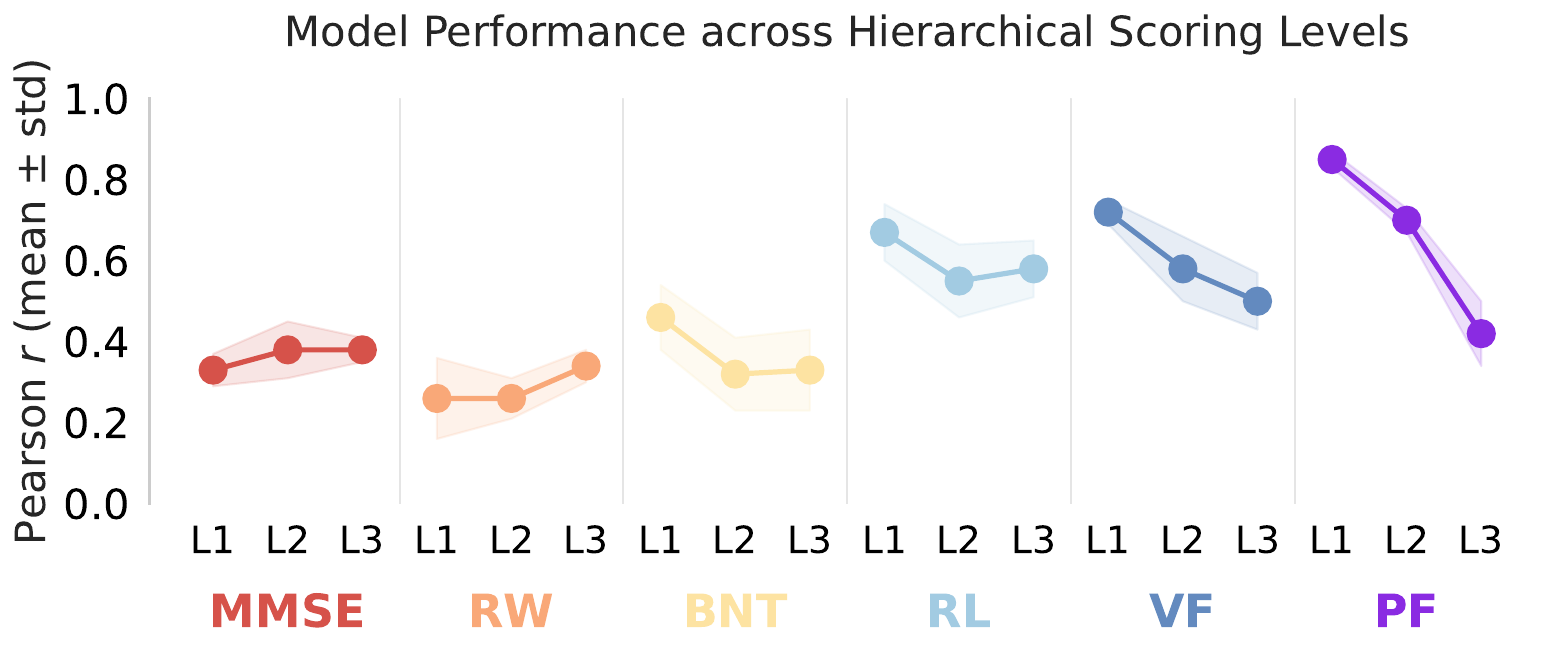}
  \caption{Hierarchical prediction patterns across aggregation levels. Lines depict mean Pearson correlation $r$ across 5-fold cross-validation for each task at Level~1 (Individual Tests), Level~2 (Cognitive Domains), and \textbf{Level~3 (CERAD+ total)}.}
  \label{fig:line_plot}
\end{figure}

\subsection{Level 3: Global Status Score Predictions}
Extending the domain-level analysis to continous global cognition score (CERAD+ total), task grouping reveals distinct aggregation dynamics (Fig.~\ref{fig:line_plot}). Open-ended tasks (PF, VF) exhibit dilution, with predictive performance decreasing from task (L1) to global-level (L3) scores. RL shows relatively stable performance across aggregation levels. In contrast, constrained tasks (MMSE, RW) show inverse dilution, where aggregation improves prediction. 
\subsection{Generalizability to Hold-out \& Feature Importance}
\begin{figure}[t]
  \centering
  \includegraphics[width=\linewidth]{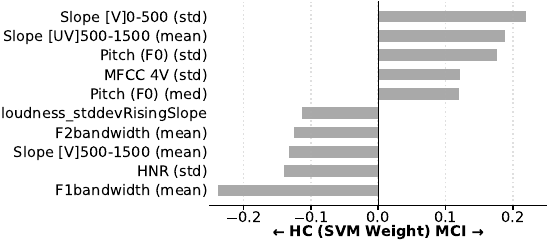}
  \caption{Feature Importance of best \textbf{Level 3: MCI Binary} Prediction Model (SVM Weight-Based)}
  \label{fig:feature_importance}
\end{figure}

\begin{table}[t]
\caption{Top-performing models across the scoring hierarchy, reporting balanced accuracy (Binary) and Pearson correlation on the Development and disjoint Hold-out sets.}
\label{tab:top_models_summary}
\centering
\scriptsize
\setlength{\tabcolsep}{3pt}
\renewcommand{\arraystretch}{1.0}

\begin{tabular}{lcccc}
\toprule
\textbf{Level -- Target} & \textbf{Input Test} & \textbf{Feature} & \textbf{DEV Set} & \textbf{HO Set} \\
\midrule
Level 3: MCI (Binary)   & MMSE & eGeMAPS All & 0.62 $\pm$ 0.07 & 0.63 \\
Level 3: CERAD+ (Binary)& RL   & HuBERT      & 0.70 $\pm$ 0.01 & 0.65 \\
\midrule
Level 3: CERAD+ (Total) & RL   & HuBERT      & 0.58 $\pm$ 0.07 & 0.49 \\
Level 2: LAN            & PF   & HuBERT      & 0.70 $\pm$ 0.03 & 0.68 \\
Level 1: PF             & PF   & HuBERT      & 0.85 $\pm$ 0.02 & 0.80 \\
\bottomrule
\end{tabular}
\end{table}
Following hierarchical analyses, we evaluated model generalizability on an independent hold-out (HO) set and examined feature importance for binary MCI classification. Tab.~\ref{tab:top_models_summary} shows that HO performance closely matches the mean performance on the development (DEV) set across hierarchical levels, indicating robust generalization. HuBERT achieves the strongest performance for continuous targets and binarized CERAD+ scores. In contrast, MCI classification performs best using MMSE recordings with eGeMAPS (DEV:~$0.62 \pm 0.07$; HO:~$0.63$). Feature importance derived from SVM weights for this model on the HO set highlights interpretable acoustic correlates of impairment (Fig.~\ref{fig:feature_importance}). Positive coefficients (associated with MCI) include increased low-frequency spectral slope variability ($+0.22$) and elevated $F_0$ instability ($+0.18$). Moreover, HCs show wider $F_1/F_2$ bandwidths. A polarity shift is observed for spectral slope: steeper slopes in voiced segments are associated with HC, whereas steeper slopes in unvoiced segments correlate with MCI.

\section{Discussion and Future Work}
In this study, we demonstrated that the predictive performance of speech features depends heavily on both the hierarchical level of the cognitive target and the nature of the task itself. For open-ended tasks such as phonemic fluency, we observed a clear \textit{dilution effect}, with predictive performance declining at higher aggregation levels. These tasks can be viewed as ``specialist'' tasks: they are optimized to capture specific cognitive processes. However, because global cognition is multi-domain and the signal from a specialist task captures only a subset of the construct, predictive performance is diluted. 

In contrast, more constrained screening tasks (such as the MMSE and RW) appear to function as ``generalists''. These tasks exhibited an \textit{inverse dilution effect}, with predictive performance improving at higher levels of the hierarchy. Individual items often show ceiling effects, with both MCI and HC groups achieving near-perfect scores, but aggregating items across the full assessment increases predictive performance. This generalist profile is further supported by the MMSE-based models' ability to predict executive function scores derived entirely from non-speech drawing tasks, as well as language domain scores derived solely from speech. This suggests that speech features from structured screenings capture a cross-modal signature of cognitive health. Consistent with this, our MMSE-based MCI model achieves the strongest binary classification performance across tasks and relies on interpretable eGeMAPS features, specifically increased $F_0$ and spectral slope instability in the MCI group. These markers point to reduced speech motor control and phonatory instability, consistent with evidence that MCI is associated with greater acoustic instability and altered voice quality~\cite{themistocleous2020voice}. 

Despite promising results, our evaluation has limitations. It is limited to a single German-speaking cohort and omits socio-demographic and lifestyle covariates. Future work should test whether the proposed ``specialist'' and ``generalist'' profiles generalize across languages and cultural contexts. Joint hierarchical modeling may further capture dependencies between individual tests and global cognitive levels, improving the robustness and interpretability of speech-based cognitive monitoring.

\section{Generative AI Use Disclosure}
Generative AI tools were used only for minor language editing and to improve readability. All research ideas, study design, experiments, analyses, and interpretations were conceived and carried out by the authors. The authors take full responsibility for the originality, validity, and integrity of the work.
\section{Acknowledgements}
This research was funded by Gemeinnützigen Hertie-Stiftung and the Deutsche Forschungsgemein-
schaft (DFG) through RU 5187 (project number 442075332)and RU 5363 (project number 459422098). Additional support was provided by the Machine Excellence Cluster and DFG through the Germany’s Excellence Strategy (EXC 2064
- project number 390727645) and the following projects: CRC 1404 (project number 414984028) and
TRR 265 (project number 402170461). The authors gratefully acknowledge Dr. Ulrike Sünkel and Dr. Anna-Katharina von Thaler for their valuable assistance with data collection and annotation. The authors thank the International Max Planck Research School for Intelligent Systems (IMPRS-IS) for supporting Serli Kopar.
\bibliographystyle{IEEEtran}
\bibliography{mybib}
\end{document}